\documentclass[runningheads]{llncs}
\usepackage{graphicx}

\usepackage{tikz}
\usepackage{comment}
\usepackage{amsmath,amssymb} 
\usepackage{color}

\usepackage[accsupp]{axessibility}  

\usepackage{array}
\usepackage[caption=false,font=normalsize,labelfont=sf,textfont=sf]{subfig}
\usepackage{textcomp}
\usepackage{stfloats}
\usepackage{url}
\usepackage{verbatim}
\usepackage{cite}
\usepackage{mathrsfs}
\usepackage{booktabs}
\usepackage{array}
\usepackage{colortbl}
\usepackage{wrapfig}
\usepackage{multirow}
\usepackage{multicol}
\usepackage[mathscr]{euscript}
\usepackage{ulem}
\usepackage{diagbox}
\usepackage[ruled,linesnumbered]{algorithm2e}

\usepackage[misc]{ifsym}
\usepackage{bbding}

\makeatletter

\newcommand{\Rmnum}[1]{\expandafter\@slowromancap\romannumeral #1@}
\makeatother

\begin{document}

\pagestyle{headings}
\mainmatter

\title{Attention Diversification for Domain Generalization} 


\titlerunning{Attention Diversification for Domain Generalization}
\author{Rang Meng\inst{1, \star} 
\and
Xianfeng Li\inst{1, \star}
\and
Weijie Chen \inst{2, 1,}\textsuperscript{\Letter}
\and 
Shicai Yang \inst{1,}\textsuperscript{\Letter}
\and 
Jie Song \inst{2}
\and
Xinchao Wang \inst{3}
\and 
Lei Zhang \inst{4}
\and 
Mingli Song \inst{2}
\and 
Di Xie \inst{1}
\and 
Shiliang Pu \inst{1}
}

\authorrunning{R. Meng et al.}
%
\institute{Hikvision Research Institute, Hangzhou, China  
\and
Zhejiang University, Hangzhou, China
\and
National University of Singapore, Singapore
\and
Chongqing University, Chongqing, China
\\
\email{\{mengrang, lixianfeng6, chenweijie5, yangshicai, xiedi, pushiliang.hri\}@hikvision.com, \{sjie, songml\}@zju.edu.cn,\\
xinchao@nus.edu.sg, leizhang@cqu.edu.cn}
}

\maketitle
\renewcommand{\thefootnote}{}

\begin{abstract}
Convolutional neural networks (CNNs) have demonstrated gratifying results at learning discriminative features. However, when applied to unseen domains, state-of-the-art models are usually prone to errors due to domain shift. After investigating this issue from the perspective of shortcut learning, we find the devils lie in the fact that models trained on different domains merely bias to different domain-specific features yet overlook diverse task-related features. Under this guidance, a novel \textit{Attention Diversification} framework is proposed, in which Intra-Model and Inter-Model Attention Diversification Regularization are collaborated to reassign appropriate attention to diverse task-related features. Briefly, Intra-Model Attention Diversification Regularization is equipped on the high-level feature maps to achieve in-channel discrimination and cross-channel diversification via forcing different channels to pay their most salient attention to different spatial locations. Besides, Inter-Model Attention Diversification Regularization is proposed to further provide task-related attention diversification and domain-related attention suppression, which is a paradigm of ``\textit{simulate, divide and assemble}'': simulate domain shift via exploiting multiple domain-specific models, divide attention maps into task-related and domain-related groups, and assemble them within each group respectively to execute regularization. Extensive experiments and analyses are conducted on various benchmarks to demonstrate that our method achieves state-of-the-art performance over other competing methods. Code is available at \url{https://github.com/hikvision-research/DomainGeneralization}.
\end{abstract}

\section{Introduction} 
Domain is clarified as the feature space and marginal probability distribution for a specific dataset \cite{Shai2010A, Ben-David34}. And domain shift reveals the discrepancy between \footnotetext{$\star$ Equal contribution. \, \Letter \, Corresponding authors.}
\begin{wrapfigure}{r}{0.38\textwidth}
	\centering
	\includegraphics[width=0.4\columnwidth]{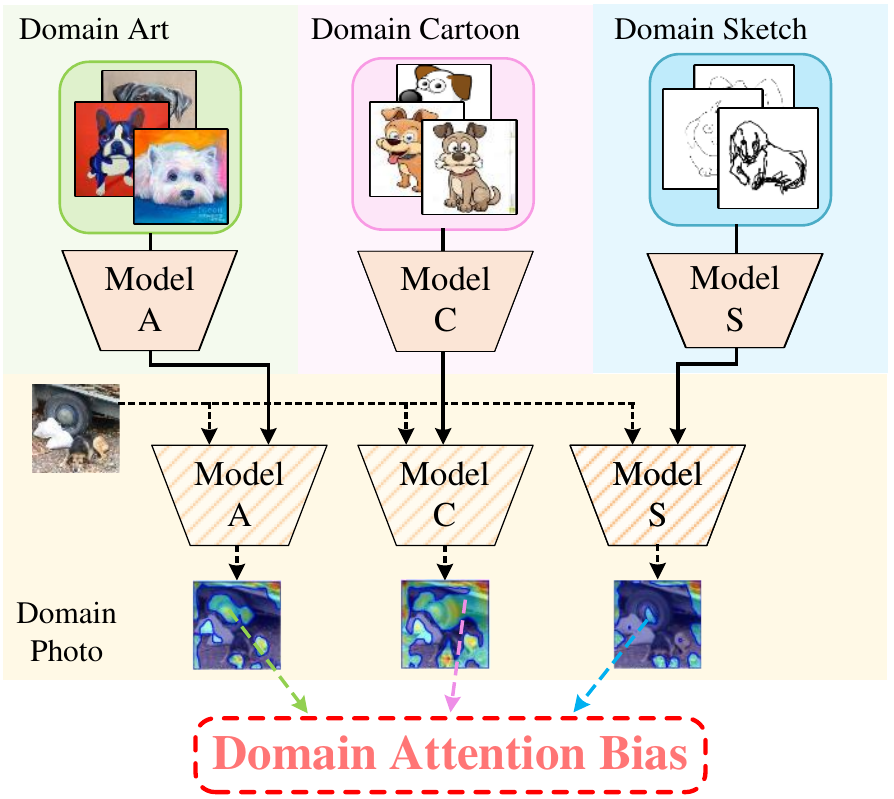} 
	\caption{The visualization of domain attention bias on PACS dataset.
	 Domain-specific models trained on different domains (ACS) pay attention to different regions when they are tested on an unseen domain (P).}
	\label{f:introduction}
\end{wrapfigure}
source and target domains~\cite{Shai2010A, Ben-David34, torralba2011unbiased}, which induces the models trained on source domains to perform defectively on an unseen target domain. Domain adaptation (DA) aims to remedy this issue of domain shift for various tasks in cases that target data is available \cite{MMD,MCD,SlimDA,chen2022learning,li2022target,yuan2022simulation,chen2021self,li2021free,zhao2022sfocda,wang2021interbn}. However, the domain shift is usually agnostic in real-world scenarios since the target data is not available for training. This issue inspires the research area of domain generalization (DG) \cite{Balaji06,muandet2013domain,zhou2022generative,zhou2022adaptive, Rahman19,Jia20,Sicilia21,li2017deeper,li2018learning,Peng37,Pan38,Zhou28,lin2021semi,sun2022dynamic,song2018selective,zhao2022shade,zhao2021m3l,li2022uncertainty}, which is aimed to make models trained on seen domains achieve accurate predictions on unseen domainss, i.e., the conditional distribution $P(Y|X)$ is robust with shifted marginal distribution $P(X)$.

Canonical DG focuses on learning a domain-invariant feature distribution $P(F(X))$ across domains for the robustness of conditional distribution $P(Y|F(X))$. 
In fact, the domain issue can be revisited from the perspective of shortcut learning \cite{Geirhos31}, which indicates that models attempt to find the simplest solution to solve a given task.
Models trained on specific domains merely pay attention to salient domain-related features while overlooking other diverse task-related information. 
When the domain shifts, the discrimination of the biased features will not be held on the unseen domain, leading to the shift of the conditional distribution. This problematic phenomenon is dubbed as ``domain attention bias'' as shown in Fig. \ref{f:introduction}.

In this paper, we propose the \textit{Attention Diversification} framework, in which the attention mechanism is served as the bridge to achieve the invariance of conditional distribution.
In our framework, the proposed Intra-Model Attention Diversification Regularization (Intra-ADR) and Inter-Model Attention Diversification Regularization (Inter-ADR) are collaborated to rearrange appropriate spatial attention to diverse task-related features from coarse to fine. The reasons why the two components are designed in our framework are detailed as follows:

\subsubsection{Intra-Model Attention Diversification Regularization.~}

According to \textit{the principle of maximum entropy} \cite{guiasu1985principle}, when estimating the probability distribution, we should select that distribution which leaves us the largest uncertainty under our constraints, so that we cannot bring any additional assumptions into our computation. That is, when testing the
unseen domains, each task-related feature is equally-useful (i.e., the maximum entropy), driving us to propose Intra-ADR, which coarsely recalls overlooked features outside the domain attention bias as much as possible. This is done via forcing different channels to pay attention to different spatial locations, leading all spatial locations to be activated. 
To this end, in-channel discrimination and cross-channel diversification are facilitated.

Although the Intra-ADR is equipped upon the high-level features,
not all spatial regions are consistent with the semantics of the categories. As stated in \cite{Geirhos31}, the background regions mainly involve domain-related features, and some parts of foreground regions are also affected by domain-specific styles \cite{Huang32,jing2020dynamic}. 
Since the Intra-ADR fails to distinguish features at the finer level into task-related and domain-related ones, the excessive attention is incidentally imposed upon domain-related features, leading to the conditional distribution shift. Thus, an attention diversification paradigm at a finer level is necessary.

\subsubsection{Inter-Model Attention Diversification Regularization.~}
To handle the aforementioned issue,  features that Intra-ADR coarsely recalls ought to be further refined by Inter-ADR. Thus, the diverse attention for task-related features is encouraged, yet the excessive attention for domain-related ones is suppressed. 
Inter-ADR is a paradigm of ``\textit{simulate, divide and assemble}''.
Specifically, 1) ``\textit{simulate}'': we train multiple domain-specific models for each seen domain, and then infer these models on samples from other training domains to simulate domain shift. In addition, the attention maps and predictions for agnostic domains are generated; 2) ``\textit{divide}'': we divide attention maps from domain-specific models and domain-aggregated model into the task-related and domain-related groups, according to whether the model predictions is consistent with the corresponding ground truth;
3) ``\textit{assemble}'': attention maps from different models are assembled within each group as the task-related and domain-related inter-model attention maps, respectively.
Finally, the attention maps of the domain-aggregated model can be regularized with the task-related and domain-related inter-model attention maps, to diversify task-related attention regions yet suppress domain-related attention regions. 

Extensive experiments and analyses are conducted on multiple domain generalization datasets. Our optimization method achieves state-of-the-art results. It is worth emphasizing that our method can bring further performance improvement in conjunction with other DG methods. 

\section{Related Works}
\noindent\textbf{Domain Generalization.~}
The analysis in~\cite{Ben-David34} proves that the features tend to be general and can be transferred to unseen domains if they are invariant across different domains. Following this research, a sequence of domain alignment methods is proposed, which reduce the feature discrepancy among multiple source domains via aligning domain-invariant features.  These methods enable models to generalize well to unseen target domains. Specifically, they use explicit feature alignment by minimizing the maximum mean discrepancy (MMD)~\cite{Tzeng35} or using Instance Normalization (IN) layers~\cite{Pan38}. Alternatively, \cite{Rahman19,Jia20} adopt domain adversarial learning for domain alignment, which trains a discriminator to distinguish the domains while training feature extractors to cheat the domain discriminator for learning domain-invariant features. 
Besides, the ability of generalizing to unseen domains will increase as training data covering more diverse domains. Several domain diversification attempts had been implemented in previous works: swapping the shape or style information of two images~\cite{Li27}, mixing instance-level features of training samples across domains~\cite{Zhou28}, altering the location and scene of objects~\cite{Prakash39}, and simulating the actual environment for generating more training data~\cite{Tobin40}. In contrast, we investigate the issue of DG inspired by shortcut learning and maximum entropy principle. Besides, we introduce visual attention in our proposed method to boost DG, which is seldom studied in prior works.

\noindent\textbf{Visual Attention.~}
Visual attention has been widely used in deep learning and achieves remarkable advances~\cite{Vaswani50,yang2020CVPR}. It has been exploited in computer vision tasks such as image recognition \cite{Wang53,Sucheng2022CVPR,Weihao22MetaFormer,chen2019all,lin2019attribute,chen2019layer} and object detection among others~\cite{YujingCVPR22,Huihui21AAAI,KehongCVPR22,XinyiECCV22,chen2022label}.
CAM \cite{2016Learning} provides the attention visualization of feature maps for model interpretable analysis. In essence, visual attention can be interpreted as an allocation mechanism for the model learning resource: it assigns high weights to what the model considers valuable, and vice versa, assigns low weight to what the model considered negligible~\cite{YidingECCV20}. Motivated by this mechanism, many computer vision tasks achieve breakthrough. For example, many fine-grained image classification methods learn multi-attention to capture sufficient subtle inter-category differences \cite{2019Looking,2020Channel,2018Multi,yang2020NeurIPS}. Recently, self-attention \cite{2017Non,0Relation,2020Dual} has emerged to model the long-range dependencies. In the field of transfer learning, Attentional Heterogeneous Transfer (AHT) \cite{Moon54} designed a new heterogeneous transfer learning approach to transfer knowledge from an optimized subset of source domain samples to a target domain. Transferable Attention for Domain Adaptation (TADA) \cite{Wang55} is proposed to use transferable global and local attention with multi-region-level domain discriminators to pick out the images and the transferable areas of the image. 

Our work finds that CNN allocates sufficient attention to domain-related features, but insufficient attention to task-related features conversely. Under this consideration, we adopt spatial attention as a bridge to learn diverse transferable features to mitigate domain shifts.

\begin{figure*}[t]
	\centering
	\includegraphics[width=1\columnwidth]{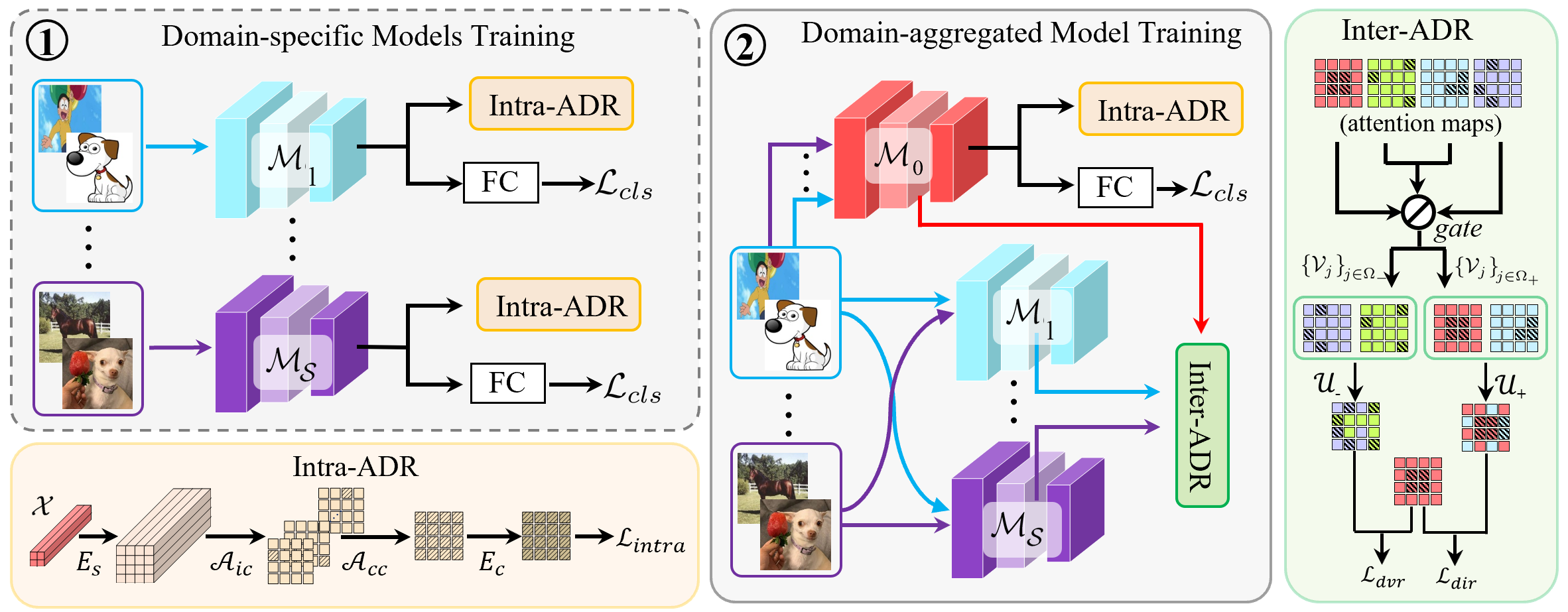} 
	\caption{The pipeline of our proposed \textit{Attention Diversification} framework, which is composed of Intra-ADR and Inter-ADR.}
	\label{f:pipeline}
\end{figure*}

\section{Method}
Our proposed \textit{Attention Diversification} framework is composed of Intra-ADR and Inter-ADR as shown in Fig. \ref{f:pipeline}. Our framework aims to deny shortcut learning, which ignores numerous task-related features. The Intra-ADR and Inter-ADR are collaborated to diversify attention regions for task-related features.\\

\noindent\textbf{Notations.~}Given $\mathcal{S}$ training domains $\{\mathcal{D}_d\}_{d=1}^{\mathcal{S}}$, where $\mathcal{D}_d=\{(x_i^d, y_i^d)\}_{i=1}^{N_d}$ with $N_d$ labeled samples covering $Z$ categories. Let $\mathcal{M}$ denote the CNN model used for image classification.
Suppose $\mathcal{X}_{j}^{b} \in R^{\mathcal{C}^b \times \mathcal{H}^b \times \mathcal{W}^b}$ denote the feature maps output from the $b$-th block of the model $\mathcal{M}_j$, where $\mathcal{C}^b$, $\mathcal{H}^b$ and $\mathcal{W}^b$ denote the channel number, height and width of $\mathcal{X}_{j}^{b}$, and $b \in \{1,...,B\}$. 
We denote the domain-specific models and domain-aggregated model as $\{\mathcal{M}_j\}_{j=0}^{\mathcal{S}}$, where $\mathcal{M}_1, ...,\mathcal{M}_{\mathcal{S}}$ represent the former which is trained on the corresponding single training domain, and $\mathcal{M}_{0}$ represents the later which is trained on multiple training domains. For the image classification task, the cross-entropy loss is employed as supervision:
\begin{equation}
\mathcal{L}_{cls} = \mathcal{L}_{CE}(\mathcal{M}_{j}(x_i^d), y_i^d)
\label{eq:cls}
\end{equation}

\subsection{Intra-Model Attention Diversification Regularization}
\label{s:intra}
In this section, we introduce the design of Intra-ADR, which forces different channels to pay their most salient attention to different spatial locations, leading all spatial locations to be activated. To this end, potential features at all spatial regions are learned as much as possible.
Intra-ADR is equipped upon the feature maps $\mathcal{X}^B$ for the last convolutional block.\\

\noindent\textbf{In-Channel Attention Map.~} 
We perform normalization to each channel in $\mathcal{X}^B$ via spatial softmax, and obtain in-channel attention maps for different channels. In doing so, the maximum of the sum of all in-channel attention maps is permanently fixed as 1:
\begin{equation}
\mathcal{A}_{ic}(\mathcal{X}^B_{c,h,w})=\frac{exp(\mathcal{X}^B_{c,h,w})}{\sum_{h=1}^{\mathcal{H}^B}\sum_{w=1}^{\mathcal{W}^B}exp(\mathcal{X}^B_{c,h,w})}
\label{eq:aic}
\end{equation}
where $\mathcal{A}_{ic}(\cdot)$ is the operation of spatial softmax, and $\mathcal{X}^B_{c,h,w}$ denotes the pixel at the spatial location $(h,w)$ of the $c$-th channel in $\mathcal{X}^B$.  In this way, when the magnitudes of the selected pixels are enhanced, pixels in the remaining spatial location will be suppressed conversely. This means that attention is concentrated on the selected pixels and then we obtain ``sparse'' in-channel attention maps. \\

\noindent\textbf{Cross-Channel Attention Map.~}
Inspired by the maxout operation in~\cite{Goodfellow43}, we enforce Pixel-wise Cross-Channel Maximization upon $\mathcal{A}_{ic}$ to obtain the cross-channel attention map:
\begin{equation}
\mathcal{A}_{cc}(\mathcal{X}_{c,h,w}^B)=\mathop{\max}_{c=1,2,...,C^B}\mathcal{A}_{ic}(\mathcal{X}^B_{c,h,w})
\label{eq:acc}
\end{equation}where $\mathcal{A}_{cc}(\cdot)$ is the Pixel-wise Cross-Channel Maximization, and $\mathcal{A}_{cc}(\mathcal{X}^B_{c,h,w}) \in \mathbb{R}^{\mathcal{H}^B \times \mathcal{W}^B}$ contains the most representative pixels across different channels.
Thus, when we maximize the sum of $\mathcal{A}_{cc}(\mathcal{X}^B_{c,h,w})$, all spatial locations are extremely activated to achieve ``\textit{dense}'' features.\\

\noindent\textbf{Spatial-Channel Joint Expanding Module.~}
However, the involving channels in the cross-channel attention map are limited because of the following observation: 
take ResNet-50 with input size of 224 \cite{He57} as an example, the number of spatial location of $\mathcal{X}^B$ ($\mathcal{H}^B \times \mathcal{W}^B=49$) is far less than that of channels ($\mathcal{C}^B=2048$). The majority of channels are not involved in regularization, leading to that lots of features cannot get sufficient attention. To remedy this issue, we propose a Spatial-Channel joint Expanding module (SCE) to enlarge both the spatial scope and involved channel number in the cross-channel attention map. SCE consists of two strategies:

\begin{itemize}
    \item \textit{Spatial Expanding.~} The spatial expanding block is composed of a deconvolutional layer, an instance normalization layer and a ReLU activation layer. This is done for two-folder reasons: i) deconvolution can enlarge the resolution of feature maps to offset the gap between channel number and spatial location number; ii) deconvolution can provide more detailed semantic clues. The output of spatial expanding block $\mathbb{X}^B$ can be expressed as:
    \begin{equation}
    \mathbb{X}^B = E_{s}(\mathcal{X}^B_{c,h,w})
    \label{eq:e}
    \end{equation}
    where $E_{s}$ is the spatial expanding block, $\mathbb{X}^B\in\mathbb{R}^{\mathcal{C}^B \times \textbf{H}^B \times\textbf{W}^B}$, $\textbf{H}^B=s*\mathcal{H}^B$ and $\textbf{W}^B=s*\mathcal{W}^B$. $s>1$ is a scale factor.
    \item \textit{Channel Expanding.~}
    We involve more channels in the cross-channel attention map via the Pixel-wise Cross-Channel Top-$k$ selection, which averages the most activated $k$ pixels across channels:
    \begin{equation}
    \mathbb{A}_{cc}(\mathbb{X}^B_{c,h,w})=E_{c}(\mathcal{A}_{ic}(\mathbb{X}^B_{c,h,w}))= \mathop{\max(k)}\limits_{c=1,2,...,\mathcal{C}^B} \mathcal{A}_{ic}(\mathbb{X}^B_{c,h,w})
    \label{eq:topk}
    \end{equation}
    where $k$ is the number of selected channels, $\mathbb{A}_{cc}(\mathbb{X}_{c,h,w})$ is the output of SCE, and $\max(k)(\cdot)$ is the operation of averaging the most activated $k$ pixels across channels. Note that SCE can make the cross-channel attention map involve $k \cdot s$ times channels compared with the original one.\\
\end{itemize}

\noindent\textbf{Intra-Model Regularization.~}
We impose SCE upon the output feature maps from the last convolutional block, and formulate the Intra-ADR term via maximizing the average value of $\mathbb{A}_{cc}(\mathbb{X}^B_{c,h,w})$:
\begin{equation}
\mathcal{L}_{intra}=-\frac{1}{\textbf{HW}}\sum_{h=1}^{\textbf{H}}\sum_{w=1}^{\textbf{W}}\mathop{\max(k)}\limits_{c=1,2,...,\mathcal{C}^B} \mathcal{A}_{ic}(\mathbb{X}^B_{c,h,w})
\label{eq:ADR}
\end{equation}
\subsection{Inter-Model Attention Diversification Regularization}
In this section, we introduce the other component of our proposed framework, i.e., Inter-ADR, which refines the attention assignment of Intra-ADR by a paradigm of ``\textit{simulate, divide and assemble}''. This is done to diversify task-related features yet suppress the domain-related features.
Details are as follows:\\

\noindent\textbf{Simulate Domain Shift.~} 
We train the domain-specific models $\{\mathcal{M}_{j}\}^\mathcal{S}_{j=1}$ on each source domain with $\mathcal{L}_{intra}$ in Eqn. \ref{eq:ADR} and $\mathcal{L}_{cls}$ in Eqn. \ref{eq:cls}. We do not only infer each sample on its own domain-specific model but also on other domain-specific models to simulate domain shift.
Then we obtain the cross-channel attention maps for the $b$-th block $\{\{\mathcal{V}_j^b\}_{b=1}^B\}_{j=0}^{\mathcal{S}}$ from each domain-specific model in the same manner of Intra-ADR (without spatial expanding block) : 
\begin{equation}
\mathcal{V}_{j}^{b}=
\mathop{\max(\mathcal{C}^b)}\limits_{c=1,2,...,\mathcal{C}^b} \mathcal{A}_{ic}(\mathcal{X}_{j}^{b})
\end{equation}

\noindent\textbf{Divide Attention Maps Across Models.~} 
The models prediction $\hat{y}_j$ is utilized as the criterion to divide cross-channel attention maps $\{\{\mathcal{V}_j^b\}_{b=1}^B\}_{j=0}^{\mathcal{S}}$ from different models. $\{\{\mathcal{V}_j^b\}_{b=1}^B\}_{j=0}^{\mathcal{S}}$ are divided into two groups: if the prediction  $\hat{y}_j$ agrees with the corresponding ground truth, $\{\mathcal{V}_j^b\}_{j=1}^B$ are viewed as task-related features, otherwise
domain-related features. Let $\{\{\mathcal{V}_j^b\}_{b=1}^B\}_{j\in \Omega_+}$ and $\{\{\mathcal{V}_j^b\}_{b=1}^B\}\}_{j\in \Omega_-}$ denote the two groups respectively:
\begin{equation}
\begin{aligned}
&j \in
&\left\{\begin{array}{ll}
\Omega_+, & \text {if} \quad \hat{y}_i^j = y_i^d\\
\Omega_-, & \text {otherwise}
\end{array}\right.
\quad s.t. \quad  j=0,..., \mathcal{S}
\end{aligned}
\end{equation}
\noindent\textbf{Assemble Attention Maps in Each Group.~} 
We assemble the cross-channel attention maps for each group via Pixel-wise Cross-Model Maximization, which is similar to Pixel-wise Cross-Channel Maximization in Eqn. \ref{eq:acc}:
\begin{equation}
\mathcal{U}_+^b = \mathop{\max}_{j \in \Omega_+} \mathcal{V}_{j}^{b}, \quad
\mathcal{U}_-^b = \mathop{\max}_{j \in \Omega_-} \mathcal{V}_{j}^{b}
\label{eq:u}
\end{equation}
where $\mathcal{U}_+^b$ and $\mathcal{U}_-^b$ are the task-related and domain-related inter-model attention map, respectively. Thanks to the Pixel-wise Cross-model Maximization, $\mathcal{U}_+^b$ contains appropriate attention regions attributed to correct predictions under domain shift. On the other hand, $\mathcal{U}_-^b$ includes the most salient attention locations, which involve the domain-related features leading to error predictions.\\

\noindent\textbf{Inter-Model Regularization.~}
After dividing and assembling the attention maps across models, we exploit the inter-model attention map to force the cross-channel attention maps $\{\mathcal{V}_{0}^b\}_{b=1}^{B}$ of the domain-aggregated model to encourage task-related features yet suppress domain-related features. 

{\centering
\begin{minipage}{.98\linewidth}

\begin{algorithm}[H]
\footnotesize

\SetKwInOut{Input}{input}
\SetKwInOut{Output}{output}
\Input{training data $\{\mathcal{D}\}_{d=1}^{\mathcal{S}}$, domain-specific models $\{\mathcal{M}_{j}\}_{j=1}^{\mathcal{S}}$, domain-aggregated model $\mathcal{M}_{0}$
}
\Output{trained domain-aggregated model $\mathcal{M}_{0}$\;}
\BlankLine

\For{\textit{d in $\{1,...,\mathcal{S}\}$}}
{
    Train domain-specific models $\mathcal{M}_{j=d}$ on the domain $\mathcal{D}_d$ with cross-entropy $\mathcal{L}_{cls}^d$ in Eqn. (1) and Intra-ADR losses
    $\mathcal{L}_{intra}^d$ in Eqn. (6) \;
}

\For{\textit{$\{x_i^d,y_i^d\}$ in $\{\mathcal{D}\}_{d=1}^{\mathcal{S}}$}}
{
    Generate predictions and cross-channel attention maps for domain-specific and domain-aggregated models: $\{\hat{y}_i^j=\mathcal{M}_{j}(x_i^d)\}^{\mathcal{S}}_{j=0}\}$;$\{\{\mathcal{V}^b_{j}\}_{b=1}^{B}\}^{\mathcal{S}}_{j=0}\}$\;
    Divide cross-channel attention maps from multiple models into two group based on whether the prediction agrees with the ground truth using Eqn. (8):
    $\{\{\mathcal{V}^b_{j}\}_{b=1}^{B}\}_{j \in \Omega_+}, \{\{\mathcal{V}^b_{j}\}_{b=1}^{B}\}_{j \in \Omega_-}\gets\{\{\mathcal{V}^b_{j}\}_{b=1}^{B}\}^{\mathcal{S}}_{j=0}\}$\;
    Generate task-related and domain-related inter-model attention maps $\{\mathcal{U}_+^b\}_{b=1}^{B}$ and $\{\mathcal{U}_-^b \}_{b=1}^{B}$ using Eqn. (9)\;
    Calculate $\mathcal{L}_{dir}$ and $\mathcal{L}_{dvr}$ using Eqn. (10) and Eqn. (11)\;
    Train domain-aggregated model $\mathcal{M}_{0}$ with $\mathcal{L}_{total}$ in Eqn. (13)
}
\caption{Attention Diversification Training Schema}
\label{algo:adr}
\end{algorithm}
\end{minipage}
}

On the one hand, we minimize the Euclidean distance between $\{\mathcal{V}_0\}_{b=1}^B$ and $\{\mathcal{U}_+^b\}_{b=1}^B$ to enhance attention regions involving task-related features:
\begin{equation}
\mathcal{L}_{dir}=\sum_{b=1}^{B}\|(\mathcal{V}_{0}^{b}-\mathcal{U}_{+}^{b})\|_{2}
\end{equation}

On the other hand, we ought to suppress the attention regions involving domain-related features,  which are accidentally included during Intra-ADR. This is done through maximizing the Euclidean distance between $\{\mathcal{V}_0\}_{b=1}^B$ and $\{\mathcal{U}_-^b\}^B_{b=1}$:
\begin{equation}
\mathcal{L}_{dvr}=-\sum_{b=1}^{B}||(\mathcal{V}_{0}^{b}-\mathcal{U}_{-}^{b})||_{2}
\end{equation}

Therefore, the Inter-ADR term can be expressed as:
\begin{equation}
\mathcal{L}_{inter}=\lambda_{dir} \cdot \mathcal{L}_{dir} +\lambda_{dvr} \cdot \mathcal{L}_{dvr}
\label{eq:ADR2}
\end{equation}
where $\lambda_{dir}$ and $\lambda_{dvr}$ are the hyperparameters to balance the two losses. \\

\noindent\textbf{Training Scheme.~}
Our framework is trained in a two-stage manner, which includes domain-specific models training and domain-aggregated-model training as shown in Algorithem \ref{algo:adr}. In the first stage, only Eqn. \ref{eq:ADR} is used for attention diversification. In the second stage, both Eqn. \ref{eq:ADR} and Eqn. \ref{eq:ADR2} are involved into attention diversification training. The total loss is:
\begin{equation}
\begin{split}
\mathcal{L}_{total}=&\mathcal{L}_{cls} +\lambda_{intra} \cdot \mathcal{L}_{intra}
+ \lambda_{dir} \cdot \mathcal{L}_{dir} +\lambda_{dvr} \cdot \mathcal{L}_{dvr}
\end{split}
\end{equation}

\section{Experiments}
In this section, we demonstrate the effectiveness of our \textit{Attention Diversification} framework on three mainstream DG benchmarks.
For convenience, we abbreviate our \textit{Attention Diversification} framework to I$^{2}$-ADR.

\subsection{Experimental Setup}
\subsubsection{Datasets.~} 
\textbf{PACS}~\cite{li2017deeper} is a common DG benchmark that contains 9991 images of 7 categories from 4 different domains, i.e., Art (A), Cartoon (C), Photo (P), and Sketch (S).
\textbf{Office-home}~\cite{Venkateswara48} contains images sharing 65 categories from 4 different domains around 15,579 images, i.e., Art (Ar), Clipart (Cl), Product (Pr), and Real-World (Rw). 
\textbf{DomainNet} \cite{Peng45} is a very large DG dataset, consisting of about 600K images with 345 categories from 6 different domains, i.e., Clipart (C), Infograph (I), Painting (P), Quickdraw (Q), Real (R), Sketch (S).

\subsubsection{Implementation Details.~}
Our framework is trained from ImageNet \cite{Deng49} pre-trained models. We utilize an SGD optimizer, batch size of 64 and weight-decay of 0.0004 with 150 epochs for optimization. The initial learning rate is set 0.008 and adjusted by a cosine annealing schedule. Following the standard augmentation protocol in~\cite{Carlucci08}, we train our framework with horizontal flipping, random cropping, color jittering, and grayscale conversion. We follow the standard data splits and leave-one-domain-out evaluation protocol as the prior work \cite{Carlucci08}. The best models are selected based on the validation split of training domains. The accuracy for test domains is reported and averaged over three runs.
We mainly use ResNet-18/50 as the backbones. Note that the same backbone is adopted among the domain-aggregated and
the domain-specific models. We set the hyperparameters, $\lambda_{intra}$, $\lambda_{dir}$, and $\lambda_{dvr}$ as $0.005$,  $2$ and $1$ for all datasets, respectively. Our framework is implemented with PyTorch on NVIDIA Tesla V100 GPUs.

\begin{table*}[t]
	\centering
 \caption{Leave-one-domain-out generalization results on PACS dataset.}
	\renewcommand{\arraystretch}{0.95}
	\setlength\tabcolsep{1.5pt}
	\resizebox{\textwidth}{!}
	{
	\begin{tabular}{l|c||cccc|c|c|cccc|c}
			
			\specialrule{0.075em}{0pt}{1pt}
			
			\multicolumn{1}{l|}{\multirow{2}{*}{\textbf{Methods}}} &\multicolumn{1}{c||}{\multirow{2}{*}{\textbf{References}}}&\textbf{Art} &\textbf{Cartoon}&\textbf{Photo}&\textbf{Sketch}&\textbf{Avg.} &&\textbf{Art} &\textbf{Cartoon}&\textbf{Photo}&\textbf{Sketch}&\textbf{Avg.}\\
			\cline{3-7} \cline{9-13} 
			&&\multicolumn{5}{c|}{\textit{ResNet-18}} &&\multicolumn{5}{c}{\textit{ResNet-50}}         \\
			\specialrule{0.05em}{1pt}{1pt}
			Baseline  &-&79.0&74.3&94.9&71.4&79.9&&86.2&78.7&97.6&70.6&83.2\\
			MetaReg~\cite{Balaji06}&NeurIPS'18&83.7&77.2&95.5&70.3  &81.7&&87.2&79.2&97.6&70.3&83.6     \\
			MASF~\cite{Dou09}&NeurIPS'19&80.2&77.1&94.9&71.6  &81.0&&82.8&80.4&95.0&72.2&82.6 \\
			
			Epi-FCR~\cite{Li56}&ICCV'19&82.1&77.0&93.9&73.0  &81.5       &&-&-&-&-&-\\
			
			JiGen~\cite{Carlucci08}&CVPR'19&79.4&75.2&96.0&71.3  &80.5   &&-&-&-&-&- \\
			DMG~\cite{Chattopadhyay46}&ECCV'20&76.9&80.4&93.4&75.2&81.5&&82.6&78.1&94.5&78.3&83.4\\		
			RSC~\cite{Huang32}&ECCV'20&84.4&80.3&95.9&80.8   &85.1       &&87.8&82.1&97.9&83.3&87.9\\ 
			MixStyle~\cite{Zhou28}&ICLR'21&84.1&78.8&\uline{96.1}&75.9&83.7     &&-&-&-&-&- \\
			SelfReg~\cite{Kim73}&ICCV'21&82.3&78.4&\textbf{96.2}&77.5&83.6&&87.9&79.4&96.8&78.3&85.6\\
			DAML~\cite{shu2021open}&CVPR'21&83.0&74.1&95.6&78.1&82.7&&-&-&-&-&-\\
			SagNet~\cite{Nam66}&CVPR'21&83.6&77.7&95.5&76.3&83.3&&81.1&75.4&95.7&77.2&82.3\\
			FACT~\cite{xu77}&CVPR'21&\uline{85.4}&78.4&95.2&79.2&84.5&&\textbf{89.6}&81.7&96.8&84.4&88.1\\
		
			\specialrule{0.05em}{1pt}{1pt}				
								
			Intra-ADR &Ours&82.4&79.4&95.3&82.3&84.9&&87.7&81.2&97.1&83.8&87.5 \\			
			I$^{2}$-ADR &Ours&82.9&\uline{80.8}&95.0&83.5&85.6&&88.5&\uline{83.2}&95.2&\textbf{85.8}&88.2 \\
			MixStyle + Intra-ADR&Ours&\textbf{86.0}&80.3&96.0&\uline{84.4}&\uline{86.7}&&\uline{88.6}&\uline{83.2}&\uline{98.0}&85.2&\uline{88.7}\\
			MixStyle + I$^{2}$-ADR &Ours&85.3&\textbf{81.2}&95.4&\textbf{86.1}&\textbf{87.0}&&87.7&\textbf{84.5}&\textbf{98.2}&\uline{85.6}&\textbf{89.2}\\
			
			\specialrule{0.05em}{1pt}{1pt}		
	\end{tabular}}	
	\label{t:PACS}	
\end{table*}

\subsection{Comparison with State-of-The-Arts}

\subsubsection{Results on PACS.~}
Our framework achieves SOTA results on PACS dataset with both ResNet-18 and ResNet-50.
In Table \ref{t:PACS}, the average performance of our framework achieves 85.6\% and 88.2\% with ResNet-18 and ResNet-50, respectively. Our framework provides impressive improvements of 5.7\% and 5.0\% compared with the corresponding baselines. Compared with other SOTA results, our framework surpasses other competing DG methods. Note that one component of our framework for recalling overlooked features in shortcut learning, Intra-ADR, can be surprisingly superior to most of DG methods. Moreover, Inter-ADR can further lift the DG performance of Intra-ADR and other competing DG methods.

\begin{table}[t]
\centering
    \begin{minipage}[t]{0.49\textwidth}
    \centering
    \renewcommand{\arraystretch}{0.9}
    \makeatletter\def\@captype{table}\makeatother\caption{Office-Home under ResNet-18.}
    \resizebox{0.95\textwidth}{!}{\begin{tabular}{l|cccc|c}
			\specialrule{0.075em}{0pt}{1pt}
			\textbf{Methods}&\textbf{Ar} &\textbf{Cl}&\textbf{Pr}&\textbf{Rw}&\textbf{Avg.} \\
			\specialrule{0.05em}{1pt}{1pt}
			\multicolumn{6}{c}{\textit{ResNet-18}}\\
			\specialrule{0.05em}{1pt}{1pt}
			Baseline  &57.8&52.7&73.5&74.8&64.7    \\
			
			RSC~\cite{Huang32}&58.4&47.9&71.6&74.5&63.1      \\ 
			
			MixStyle~\cite{Zhou28}&58.7&53.4&74.2&\uline{75.9}&65.5 \\
			
			SagNet~\cite{Nam66}&60.2&45.4&70.4&73.4&62.3\\
			FACT~\cite{xu77}&60.3&54.9&74.5&\textbf{76.6}&66.6\\

			\specialrule{0.05em}{1pt}{1pt}

			Intra-ADR&64.5&54.0&73.9&74.7&66.8  \\
			I$^{2}$-ADR &\uline{66.4}&53.3&\uline{74.9}&75.3&67.5  \\
			MixStyle + Intra-ADR&65.9&\uline{55.3}&74.3&75.1&\uline{67.7}\\
			MixStyle + I$^{2}$-ADR&\textbf{66.8}&\textbf{56.8}&\textbf{75.3}&75.7&\textbf{68.7}\\
			
			\specialrule{0.075em}{1pt}{0pt}
	\end{tabular}}
	
	\label{t:OF18}
        \end{minipage}
        \begin{minipage}[t]{0.49\textwidth}
        \renewcommand{\arraystretch}{0.9}
        \makeatletter\def\@captype{table}\makeatother\caption{ Office-Home under ResNet-50.}
        \centering
            \resizebox{0.95\textwidth}{!}{\begin{tabular}{l|cccc|c}
			\specialrule{0.075em}{0pt}{1pt}
			\textbf{Methods} &\textbf{Ar} &\textbf{Cl}&\textbf{Pr}&\textbf{Rw}&\textbf{Avg.} \\
			\specialrule{0.05em}{1pt}{1pt}
			\multicolumn{6}{c}{\textit{ResNet-50}}\\
			\specialrule{0.05em}{1pt}{1pt}
			Baseline  &61.3&52.4&75.8&76.6&66.5\\
			MLDG~\cite{li2018learning}&61.5&53.2&75.0&77.5&66.8\\
			
			RSC~\cite{Huang32}&50.7&51.4&74.8&75.1&65.5\\
			
			SelfReg~\cite{Kim73}&63.6&53.1&76.9&78.1&67.9\\
			SagNet~\cite{Nam66}&63.4&54.8&75.8&78.3&68.1\\
			
			\specialrule{0.05em}{1pt}{1pt}
			Intra-ADR&67.3&54.1&78.8&78.8&69.8\\
			I$^{2}$-ADR&\uline{70.3}&55.1&\uline{80.7}&79.2&\uline{71.4}\\
			MixStyle + Intra-ADR&69.5&\uline{55.9}&80.6&\uline{80.4}&\uline{71.4}\\
			MixStyle + I$^{2}$-ADR&\textbf{71.1}&\textbf{56.9}&\textbf{81.8}&\textbf{80.5}&\textbf{72.5}\\
			
			\specialrule{0.075em}{1pt}{0pt}
			
	\end{tabular}}
	\label{t:OF50}
        \end{minipage}
    \end{table}

\subsubsection{Results on Office-Home.~}
From Table \ref{t:OF18} and Table~\ref{t:OF50}, it can be observed that the baseline has a strong performance on Office-Home. Many previous DG methods cannot improve or perform worse than the baseline. Nevertheless, our framework achieves 67.5\% and 71.4\% with ResNet-18 and ResNet-50, respectively. Moreover, the proposed I$^2$-ADR surpasses the majority of other related methods, including the latest MixStyle \cite{Zhou28}, RSC~\cite{Huang32}, and FACT \cite{xu77}. Notably, the Intra-ADR can achieve SOTA results of 66.8\% and 69.8\% on Office-Home with  ResNet-18 and ResNet-50, respectively.
Results on Office-Home justify the impact of each component of our framework.

\begin{table}[t]
\small
	\centering
 \caption{Leave-one-domain-out generalization results on DomainNet dataset.}
	\renewcommand{\arraystretch}{0.9}
	\setlength\tabcolsep{3pt}
	\resizebox{0.98\textwidth}{!}{\begin{tabular}{l|c|cccccc|c}
			\specialrule{0.075em}{0pt}{1pt}
			\textbf{Methods} &\textbf{References}&\textbf{Clipart} &\textbf{Infograph}&\textbf{Painting}&\textbf{Quickdraw}&\textbf{Real}&\textbf{Sketch}&\textbf{Avg.} \\
			\specialrule{0.05em}{1pt}{1pt}
			\multicolumn{9}{c}{\textit{ResNet-18}}\\
			\specialrule{0.05em}{1pt}{1pt}
			Baseline &-&57.1&17.6&43.2&\uline{13.8}&\uline{54.9}&39.4&37.6\\
			MetaReg~\cite{Balaji06}&NeurIPS'18&53.7&\textbf{21.1}&\textbf{45.3}&10.6&\textbf{58.5}&42.3&38.6\\
			DMG~\cite{Chattopadhyay46}&ECCV'20&\textbf{60.1}&\uline{18.8}&44.5&\textbf{14.2}&54.7&41.7&\textbf{39.0}\\
			
			\specialrule{0.05em}{1pt}{1pt}
			Intra-ADR&Ours&57.3$\pm$0.1&14.9$\pm$0.3&42.8$\pm$0.2&12.2$\pm$0.4&52.9$\pm$0.5&46.0$\pm$0.2&37.7\\
			I$^{2}$-ADR&Ours&57.3$\pm$0.3&15.2$\pm$0.3&44.1$\pm$0.1&12.1$\pm$0.4&53.9$\pm$0.6&\uline{46.7$\pm$0.2}&38.2\\ 
			MixStyle + Intra-ADR&Ours&\uline{57.4$\pm$0.2}&15.3$\pm$0.1&43.3$\pm$0.2&12.3$\pm$0.4&53.5$\pm$0.3&46.5$\pm$0.2&38.1\\
			MixStyle + I$^{2}$-ADR&Ours&\uline{57.4$\pm$0.4}&15.7$\pm$0.2&\uline{44.7$\pm$0.1}&12.3$\pm$0.4&54.4$\pm$0.2&\textbf{47.4$\pm$0.1}&\uline{38.7}\\
			\specialrule{0.05em}{1pt}{1pt}
			\multicolumn{9}{c}{\textit{ResNet-50}}\\
			\specialrule{0.05em}{1pt}{1pt}
			Baseline &-&62.2&19.9&45.5&13.8&57.5&44.4&40.5      \\
			MetaReg~\cite{Balaji06}&NeurIPS'18&59.8&\textbf{25.6}&\textbf{50.2}&11.5&\textbf{64.6}&50.1&43.6\\
			MLDG~\cite{li2018learning}&AAAI'18&59.1$\pm$0.2&19.1$\pm$0.3&45.8$\pm$0.7&13.4$\pm$0.3&59.6$\pm$0.2&50.2$\pm$0.4&41.2 \\
			
			C-DANN~\cite{Li67}&ECCV'18&54.6$\pm$0.4&17.3$\pm$0.1&43.7$\pm$0.9&12.1$\pm$0.7&56.2$\pm$0.4&45.9$\pm$0.5&38.3 \\
			RSC~\cite{Huang32}&ECCV'20&55.0$\pm$1.2&18.3$\pm$0.5&44.4$\pm$0.6&12.2$\pm$0.2&55.7$\pm$0.7&47.8$\pm$0.9&38.9 \\
			DMG~\cite{Chattopadhyay46}&ECCV'20&\textbf{65.2}&\uline{22.2}&\uline{50.0}&\textbf{15.7}&59.6&49.0&43.6\\
			SagNet~\cite{Nam66}&CVPR'21&57.7$\pm$0.3&19.0$\pm$0.2&45.3$\pm$0.3&12.7$\pm$0.5&58.1$\pm$0.5&48.8$\pm$0.2&40.3 \\
			
			SelfReg~\cite{Kim73}&ICCV'21&60.7$\pm$0.1&21.6$\pm$0.1&49.4$\pm$0.2&12.7$\pm$0.1&60.7$\pm$0.1&51.7$\pm$0.1&42.8\\
			
			\specialrule{0.05em}{1pt}{1pt}
			Intra-ADR&Ours&63.6$\pm$0.1&20.0$\pm$0.1&49.4$\pm$0.1&14.8$\pm$0.3&60.0$\pm$0.4&\textbf{54.4$\pm$0.1}&43.7\\
			I$^{2}$-ADR&Ours&\uline{64.4$\pm$0.2}&20.2$\pm$0.6&49.2$\pm$0.5&15.0$\pm$0.2&\uline{61.6$\pm$0.4}&53.3$\pm$0.1&\uline{44.0}      \\ 
			MixStyle + Intra-ADR&Ours&63.9$\pm$0.1&20.1$\pm$0.5&49.4$\pm$0.2&15.0$\pm$0.4&60.4$\pm$0.3&\textbf{54.4$\pm$0.1}&43.9\\
			MixStyle + I$^{2}$-ADR&Ours&64.1$\pm$0.1&20.4$\pm$0.2&49.2$\pm$0.4&\uline{15.1$\pm$0.2}&61.3$\pm$0.4&\uline{54.3$\pm$0.4}&\textbf{44.1}\\
			 \specialrule{0.075em}{1pt}{0pt}
	\end{tabular}}
	\label{t:DN50}
\end{table}

\begin{table}[t]
\centering
    \begin{minipage}[t]{0.49\textwidth}
    \centering
    \renewcommand{\arraystretch}{0.9}
    \makeatletter\def\@captype{table}\makeatother\caption{Ablation studies on the three components contained in I$^{2}$-ADR.}
    \resizebox{0.95\textwidth}{!}{\begin{tabular}{c|c|cc|cccc|c}		
			\specialrule{0.075em}{0pt}{1pt}		
			Method&\textbf{$\mathcal{L}_{intra}$}&\textbf{$\mathcal{L}_{dir}$}&\textbf{$\mathcal{L}_{dvr}$}&\textbf{Art} &\textbf{Cartoon}&\textbf{Photo}&\textbf{Sketch}&\textbf{Avg.} \\		
			\specialrule{0.05em}{1pt}{1pt}		
			\multicolumn{1}{c|}{\multirow{4}{*}{\begin{tabular}[c]{@{}c@{}}I$^{2}$-ADR\end{tabular}}}&$\checkmark$&$\textbf{-}$&$\textbf{-}$&82.4 &79.4 &\textbf{95.3} &82.3& 84.9\\
			&$\textbf{-}$&$\checkmark$ &$\checkmark$&82.3&80.0&\uline{95.1}&82.6&85.0       \\  
			&$\checkmark$&$\checkmark$&$\textbf{-}$&\uline{82.7}&\uline{80.5}&95.0&\uline{83.2}&\uline{85.4}      \\
			&$\checkmark$&$\textbf{-}$&$\checkmark$&82.5&80.2&\uline{95.1}&82.9&85.2       \\
			&$\checkmark$&$\checkmark$ &$\checkmark$&\textbf{82.9}&\textbf{80.8}&95.0&\textbf{83.5}&\textbf{85.6}       \\  		
			\specialrule{0.075em}{1pt}{0pt}		
	\end{tabular}}
	\label{t:inter-losses}
        \end{minipage}
        \begin{minipage}[t]{0.49\textwidth}
        \renewcommand{\arraystretch}{0.9}
        \centering
        \makeatletter\def\@captype{table}\makeatother\caption{Ablation studies on two strategies in SCE module.}
        \resizebox{0.95\textwidth}{!}{\begin{tabular}{c|c|c|cccc|c}		
			\specialrule{0.075em}{0pt}{1pt}		
			Method&\textbf{$E_{s}$}&\textbf{$E_{c}$}&\textbf{Art} &\textbf{Cartoon}&\textbf{Photo}&\textbf{Sketch}&\textbf{Avg.}\\		
			\specialrule{0.05em}{1pt}{1pt}		
			\multicolumn{1}{c|}{\multirow{4}{*}{\begin{tabular}[c]{@{}c@{}}Intra-ADR\end{tabular}}}&$\textbf{-}$&$\textbf{-}$&81.3&77.3&94.7&78.8&83.0\\	
			&$\textbf{-}$ &$\checkmark$&80.0&77.2&\textbf{96.0}&\uline{80.9}&83.5\\	
			&$\checkmark$&$\textbf{-}$&\uline{81.9}&\uline{79.3}&\uline{95.5}&79.3&\uline{84.0}  \\
			&$\checkmark$&$\checkmark$&\textbf{82.4}&\textbf{79.4}&95.3&\textbf{82.3}&\textbf{84.9}\\  		
			\specialrule{0.075em}{1pt}{0pt}		
	\end{tabular}}
	\label{t:adr}
        \end{minipage}
    \end{table}

\subsubsection{Results on DomainNet.~}
DomainNet is a very challenging large-scale dataset. The comparisons between our framework and other DG methods are reported in Table~\ref{t:DN50}. 
The number of data in DomainNet is much larger than other DG benchmarks, leading to be very challenging to use ResNet-18 as the backbone. Fortunately, our framework using ResNet-18 achieves competing results on DomainNet.
In addition, the performance of Intra-ADR and I$^2$-ADR using ResNet-50 are among the top ones.
We notice that the performance of our framework exceeds that of SelfReg~\cite{Kim73} by 1.6\% and DMG~\cite{Chattopadhyay46} by 0.4\%, respectively. This again verifies the superiority of our framework. 

\subsection{Ablation Studies}
In this section,  we carry out various ablation studies to dissect the effectiveness of our proposed \textit{Attention Diversification} framework. All ablation studies are conducted on PACS dataset with ResNet-18.

\subsubsection{Analysis for SCE Module.~} 
We conduct ablation studies on SCE module to analyze the effectiveness of each component in SCE module. There are two critical strategies, Spatial Expanding ($E_{s}$) and Channel Expanding ($E_{c}$) designed to facilitate the effectiveness of the Intra-ADR on the high-level features.  As shown in Table \ref{t:adr}, we can observe that both $E_{s}$ and $E_{c}$ can improve the ability of generalization across domains, and the gains of $E_{s}$ are slightly better than $E_{c}$. Note that the performance w/ $E_{c}$ on (A, C) is indeed worse than that w/o $E_{c}$ (baseline Intra-ADR), but the performance w/ $E_{c}$ is better than the baseline on average. On the other hand, we analyze the effect of the two crucial hyperparameters in SCE Module, the scale factor $s$ in $E_{s}$ and the selected channels number $k$ in $E_{c}$. As shown in Table \ref{t:s-k}, the average performance increases as the channel number is expanded. Besides, The scale factor also has a significant impact on the effectiveness of Intra-ADR. We set $k=10$, $s=2$ as the default setting for all experiments.

\subsubsection{Analysis for Different Losses in I$^{2}$-ADR.~}
We conduct ablation studies to investigate the effectiveness of
the three losses in I$^{2}$-ADR. As shown in Table \ref{t:inter-losses},  the first row denotes the results of Intra-ADR, the second row denotes the results of Inter-ADR. We can observe that $\mathcal{L}_{dir}$ + $\mathcal{L}_{dvr}$ contributes to the impressive improvement of 0.7\% on the average performance compared with Intra-ADR. After removing $\mathcal{L}_{dir}$ which diversifies task-related attention regions,
there is a drop of 0.4\% compared with $\mathcal{L}_{dir}$ + $\mathcal{L}_{dvr}$, but still an improvement of 0.3\% compared with Intra-ADR.
Besides, after removing $\mathcal{L}_{dvr}$ that suppresses the domain-related attention regions, $\mathcal{L}_{dir}$ solely surpasses the Intra-ADR by 0.5\%, but losses 0.2\% compared with $\mathcal{L}_{dir}$ + $\mathcal{L}_{dvr}$.

\begin{table}[t]
\centering
    \begin{minipage}[t]{0.49\textwidth}
    \centering
    \renewcommand{\arraystretch}{1}
    \makeatletter\def\@captype{table}\makeatother\caption{Ablation studies on the equipped positions of Intra-ADR.}
    \resizebox{0.95\textwidth}{!}{\begin{tabular}{l|cccc|c}		
			\specialrule{0.075em}{0pt}{1pt}		
			\textbf{Methods} &\textbf{Art}&\textbf{Cartoon}&\textbf{Photo}&\textbf{Sketch}&\textbf{Avg.} \\
			\specialrule{0.05em}{1pt}{1pt}
			Intra-ADR (res1) &80.8&78.4&94.7&80.1&83.5\\
			Intra-ADR (res2) &81.0&78.6&94.9&80.7&83.8\\
			Intra-ADR (res3) &\uline{81.8}&\uline{78.8}&\uline{95.5}&\uline{81.1}&\uline{84.3}\\
			Intra-ADR (res4) &\textbf{82.4}&\textbf{79.4}&\textbf{95.3}&\textbf{82.3}&\textbf{84.9}\\
			\specialrule{0.075em}{1pt}{0pt}		
	\end{tabular}}	
	\label{t:where-intra}
        \end{minipage}
        \begin{minipage}[t]{0.49\textwidth}
        \renewcommand{\arraystretch}{0.95}
        \centering
        \makeatletter\def\@captype{table}\makeatother\caption{Ablation studies on the equipped positions of Inter-ADR.}
            \resizebox{0.95\textwidth}{!}{\begin{tabular}{l|cccc|c}		
			\specialrule{0.075em}{0pt}{1pt}		
			\textbf{Methods} &\textbf{Art}&\textbf{Cartoon}&\textbf{Photo}&\textbf{Sketch}&\textbf{Avg.} \\
			\specialrule{0.05em}{1pt}{1pt}
			Intra-ADR &82.4&79.4&\uline{95.3}&82.3&84.9\\
			+ Inter-ADR (res1) &\textbf{83.0}&79.3&\textbf{95.5}&82.9&85.2\\
			+ Inter-ADR (res12) &82.6&80.0&\uline{95.3}&83.0&85.2\\
			+ Inter-ADR (res123) &82.7&\uline{80.5}&95.1&\uline{83.2}&\uline{85.4}\\
			+ Inter-ADR (res1234) &\uline{82.9}&\textbf{80.8}&95.0&\textbf{83.5}&\textbf{85.6}\\
			\specialrule{0.075em}{1pt}{0pt}		
	\end{tabular}}
	\label{t:where-inter}
        \end{minipage}
\end{table}

\begin{table}[t]
\centering
    \begin{minipage}[t]{0.49\textwidth}
    \centering
    \makeatletter\def\@captype{table}\makeatother\caption{The Seen Domain performance of our proposed method.}
	\renewcommand{\arraystretch}{0.95}
        \resizebox{0.95\textwidth}{!}{\begin{tabular}{l|c|cccc}		
    			\specialrule{0.075em}{0pt}{1pt}	
    			\textbf{Methods}&Backbone &\textbf{C\&P\&S}&\textbf{A\&P\&S}&\textbf{A\&C\&S}&\textbf{A\&C\&P} \\
    			\specialrule{0.05em}{1pt}{1pt}
    			Baseline &\multicolumn{1}{c|}{\multirow{3}{*}{ResNet-18}}&\textbf{96.4}&95.6&95.0&95.6\\
    			Intra-ADR&&96.2&\textbf{96.5}&94.9&96.3\\
    			I$^{2}$-ADR& &96.3&96.3&\textbf{95.1}&\textbf{96.6}\\
    			\specialrule{0.05em}{1pt}{1pt}
    			
    			Baseline &\multicolumn{1}{c|}{\multirow{3}{*}{ResNet-50}}&96.9&\textbf{96.8}&95.6&97.3\\
    			Intra-ADR&&97.4&96.7&95.5&\textbf{97.4}\\
    			I$^{2}$-ADR& &\textbf{97.6}&96.7&\textbf{97.2}&97.1\\
    						\specialrule{0.05em}{1pt}{0pt}		
    	\end{tabular}}
    	\label{t:seen}
        \end{minipage}
        \begin{minipage}[t]{0.5\textwidth}
        \centering
	\renewcommand{\arraystretch}{.95}
 \caption{Ablation studies on the scale factor $s$ and selected channels number $k$.}
    \resizebox{0.75\textwidth}{!}{\begin{tabular}{c|cccc|c}	
			\specialrule{0.075em}{0pt}{1pt}
			 ($k$, s)&\textbf{Art} &\textbf{Cartoon}& \textbf{Photo}&\textbf{Sketch}&\textbf{Avg.}\\
			 \specialrule{0.075em}{0pt}{1pt}
			 
			 (2, 2)&82.0&78.1&95.1&81.0&84.1\\
			 (2, 4)&\uline{82.2}&77.9&\textbf{96.0}&81.0&84.3\\
			 (4, 2)&82.1&78.2&95.0&81.2&84.2\\
			 (4, 4)&\uline{82.4}&77.9&\textbf{96.0}&81.0&84.5\\
			 
			 (10, 2)&\textbf{82.4}&\textbf{79.4}&95.3&\textbf{82.3}&\textbf{84.9}\\
			 (10, 4)&\textbf{82.4}&\uline{79.2}&\uline{95.6}&\uline{82.2}&\textbf{84.9}\\
			\specialrule{0.075em}{1pt}{0pt}
	\end{tabular}}
	\label{t:s-k}
        \end{minipage}
\end{table}

\subsubsection{Analysis for The Positions of Intra-ADR and Inter-ADR.~}
Extensive works have been discussed that different layers of CNNs have different effects on the information flow \cite{meng2020neural,matthew2014visualizing,tishby2015deep}. Here we also analyze the equipped positions of the proposed two modules, Intra-ADR and Inter-ADR. Let's denote 4 bottleneck stages of a standard ResNet backbone as \textit{res1-4}. For instance, \textit{res1} means the outputted feature maps of the first bottleneck stage.
As shown in Table \ref{t:where-intra}, Intra-ADR is limited when equipped upon the low- and middle-level feature maps, but provides an impressive improvement when upon the high-level. Besides, the hierarchical Inter-ADR achieves a significant impact on the average performance. Thereby, Intra-ADR is ewuipped on the highest layer to diversify task-related features instead of introducing too many domain-related features. Inter-ADR is equipped upon multi-level layers to facilitate information flow with a mechanism of distinguishing the task- and domain-related features.

\subsection{Discussions and Visualization}
\subsubsection{Performance on Seen Domains.~}
In this section, we report the performance of our framework on the seen domains from PACS dataset. As shown in Table \ref{t:seen}, ``A'',``C'',``P'',``S'' in the first row represent the classification accuracy on seen domains, including Art, Cartoon, Photo and Sketch, respectively. The performance of our framework, whether Intra-ADR or I$^{2}$-ADR, surpasses that of the baseline on almost all sub-tasks using ResNet-18 and ResNet-50, respectively. This verifies that our framework also improves the in-domain generalization.

\begin{table*}[!t]
\small
	\centering	
 \caption{Performance comparison on single-source DG. We train our methods with a single source domain and evaluate with other remaining target domains.}	
	\renewcommand{\arraystretch}{1} 
	\resizebox{1\textwidth}{!}{\begin{tabular}{c|cccc|c|cccc|c|cccc|c|cccc|c}
	\specialrule{0.075em}{0pt}{1pt}	
	&\multicolumn{20}{c}{Target Domain}\\
	 \specialrule{0.05em}{1pt}{1pt}
     Source& \multicolumn{5}{c|}{\textbf{Baseline}}& \multicolumn{5}{c|}{\textbf{RSC}~\cite{Huang32} (ECCV'20)}& \multicolumn{5}{c|}{\textbf{SelfReg}~\cite{Kim73} (ICCV'21)} & \multicolumn{5}{c}{\textbf{Intra-ADR} (Ours)} \\\cline{2-21}
    
     Domain& \multicolumn{1}{c}{\scriptsize{A}} & \multicolumn{1}{c}{\scriptsize{C}} &\multicolumn{1}{c}{\scriptsize{P}} & \multicolumn{1}{c|}{\scriptsize{S}} &\multicolumn{1}{c|}{Avg.} & \multicolumn{1}{c}{\scriptsize{A}} & \multicolumn{1}{c}{\scriptsize{C}} &\multicolumn{1}{c}{\scriptsize{P}} & \multicolumn{1}{c|}{\scriptsize{S}} &\multicolumn{1}{c|}{Avg.} &\multicolumn{1}{c}{\scriptsize{A}} & \multicolumn{1}{c}{\scriptsize{C}} &\multicolumn{1}{c}{\scriptsize{P}} & \multicolumn{1}{c|}{\scriptsize{S}} &\multicolumn{1}{c|}{Avg.}&\multicolumn{1}{c}{\scriptsize{A}} & \multicolumn{1}{c}{\scriptsize{C}} &\multicolumn{1}{c}{\scriptsize{P}} & \multicolumn{1}{c|}{\scriptsize{S}} &\multicolumn{1}{c}{Avg.} \\
     \specialrule{0.05em}{1pt}{1pt}
A&-&61.3&96.1&52.3&69.9&-&62.5&96.3&53.2&70.7&-&65.2&96.6&55.9&\uline{72.6}&-&64.8&94.4&64.3&\textbf{74.5}\\
C&64.1&-&81.8&75.8&73.9&69.0&-&85.9&70.4&\uline{75.1}&72.1&-&87.5&70.1&\textbf{76.6}&66.7&-&83.8&74.9&\uline{75.1}\\
P&66.1&29.5&-&32.3&41.6&66.3&26.5&-&32.1&41.6&67.7&29.0&-&33.7&\uline{43.5}&67.8&40.3&-&39.5&\textbf{49.2}\\
S&38.6&60.5&48.0&-&\uline{48.3}&38.0&56.4&47.4&-&47.3&37.2&54.0&46.1&-&45.8&42.7&61.5&46.6&-&\textbf{50.3}\\
\specialrule{0.05em}{1pt}{1pt}
Avg.&56.2&\uline{50.4}&74.6&\uline{53.4}&58.6&57.8&48.5&\uline{76.5}&51.9&58.7&\uline{59.0}&47.4&\textbf{76.7}&53.3&\uline{59.6}&\textbf{59.1}&\textbf{55.5}&74.9&\textbf{59.6}&\textbf{62.3}\\     
\specialrule{0.075em}{1pt}{0pt}		
\end{tabular}}	
	\label{t:single}
\end{table*}

\begin{figure}[t]
\centering
\includegraphics[width=.98\linewidth]{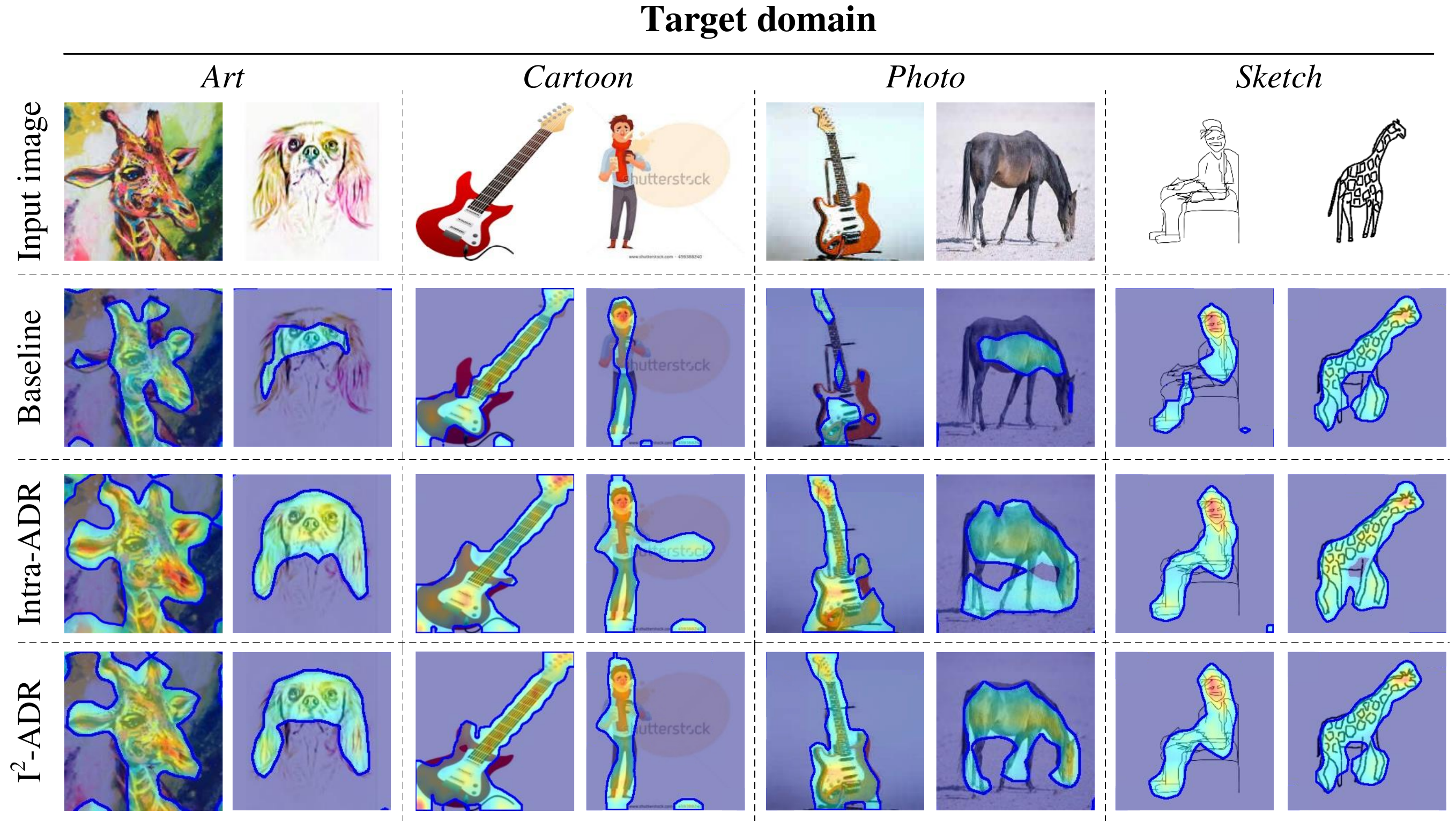}
\caption{Attention visualization on the testing domains of PACS with ResNet-18.}
\label{f:act}
\end{figure}

\subsubsection{Single-Source Domain Generalization.~}
We further evaluate our framework on single-source DG tasks. Since the Inter-ADR is not suitable for single-source DG tasks, we only report the results of Intra-ADR. Results are reported as the average accuracy among single source-target pairs.
As shown in Table~\ref{t:single}, the performance of Intra-ADR is among the top ones. This indicates that the Intra-ADR can handle both the multiple-source and single-source DG tasks, and demonstrate that diverse features effectively can avoid shortcut learning.

\subsubsection{Orthogonality to Other DG Methods.~}
Our method can also boost the performance of other DG methods. As shown in Table \ref{t:PACS}-\ref{t:DN50}, a new SOTA performance is achieved by combining our framework with MixStyle \cite{Zhou28} and is superior to other competing DG works by a significant margin.

\subsubsection{Attention Visualization.~}
We visualize the attention maps to verify our motivation and the effectiveness of our framework.
The attention maps on samples from testing split of the 4 domains in PACS are shown in Fig. \ref{f:act}. The hotter colors denote the more salient attention value, while the cooler colors represent the lower value. To compare the differences of attention regions between the baseline and our framework more clearly, we retain the top normalized attention values ($>=$ 0.7).
We can see that the proposed Intra-ADR \textit{de-facto} pays sufficient attention to diverse spatial locations, including the task-related regions and some domain-related features. Fortunately, Inter-ADR can suppress the domain-related regions and enhance the task-related regions.

\subsubsection{Limitations.~}
As shown in the last row of Fig. \ref{f:act}, there still exist some risks to maintain/enhance domain-related features in some cases. Although Inter-ADR is utilized to suppress domain-related features brought by baseline and Intra-ADR, which exploits the prediction to determine task- and domain-related features, the domain-related features will be maintained/enhanced once the corresponding cross-domain prediction is consistent with the ground-truth. Nevertheless, the proposed Intra-ADR and Inter-ADR boost the DG performance on average. The existing limitations are left as the future works.

\section{Conclusion}

Investigated from the perspective of shortcut learning, the models trained on different domains will pay attention to different salient features, aka domain attention bias. However, the principle of maximum entropy hints that every task-related feature is equally-useful potentially when encountering unseen domains. This novel insight enlightens us to remedy the issue of DG via Attention Diversification, in which we organically unify the Intra-ADR and Inter-ADR into our framework: we first utilize Intra-ADR to coarsely recall task-related features in the highest layer as much as possible, and then exploit Inter-ADR to delicately distinguish domain- and task-related features in multiple intermediate layers for further suppression and enhancement, respectively.

\section*{Acknowledgements}
This work was sponsored by National Natural Science Foundation of China (62106220, U20B2066), Hikvision Open Fund (CCF-HIKVISION OF 20210002),
NUS Faculty Research Committee~(WBS: A-0009440-00-00),
and MOE Academic Research Fund (AcRF) Tier-1 FRC Research Grant of Singapore~(WBS: A-0009456-00-00).

\clearpage
\bibliographystyle{splncs04}
\bibliography{arxiv}
\end{document}